%% file: cvpr19_stereo.tex
\documentclass[10pt,twocolumn,letterpaper]{article}

\usepackage{cvpr}
\usepackage{times}
\usepackage{epsfig}
\usepackage{graphicx}
\usepackage{amsmath}
\usepackage{amssymb}
\usepackage[usenames, dvipsnames]{xcolor}
\usepackage{multirow}
\usepackage{algorithm}
\usepackage{algorithmic}
\usepackage{amsfonts}
\usepackage{mathtools,amssymb}
\usepackage{fixltx2e}
\usepackage{bbold}
\usepackage{array}

\usepackage{setspace}

\usepackage{url}

\newcommand{\fig}[1]{Fig.~#1}
\newcommand{\tab}[1]{Table~#1}

\usepackage{expl3}
\ExplSyntaxOn
\newcommand\latinabbrev[1]{
	\peek_meaning:NTF . {
		#1\@}%
	{ \peek_catcode:NTF a {
			#1.\@ }%
		{#1.\@}}} 
\ExplSyntaxOff
\def\eg{\latinabbrev{e.g}}
\def\etal{\latinabbrev{et al}}

\def\ie{\latinabbrev{i.e}}

\usepackage[pagebackref=true,breaklinks=true,letterpaper=true,colorlinks,bookmarks=false]{hyperref}

\cvprfinalcopy 

\def\cvprPaperID{3878} 

\ifcvprfinal\fi
\begin{document}  
        
\title{Triangulation Learning Network: from Monocular to Stereo 3D Object Detection}
   
\author{
	Zengyi Qin\thanks{The work was done when Zengyi Qin was an intern at MSR.}\\
	Tsinghua University\\
	{\tt\small qinzy16@mails.tsinghua.edu.cn}
	\and
	Jinglu Wang\\
	Microsoft Research\\
	{\tt\small jinglwa@microsoft.com}
	\and
	Yan Lu\\
	Microsoft Research\\
	{\tt\small yanlu@microsoft.com}
}

\maketitle

\begin{abstract}
In this paper, we study the problem of 3D object detection from stereo images, in which the key challenge is how to effectively utilize stereo information.
Different from previous methods using pixel-level depth maps, we propose employing 3D anchors to explicitly construct object-level correspondences between the regions of interest in stereo images, from which the deep neural network learns to detect and triangulate the targeted object in 3D space. We also introduce a cost-efficient channel reweighting strategy that enhances representational features and weakens noisy signals to facilitate the learning process. All of these are flexibly integrated into a solid baseline detector that uses monocular images. We demonstrate that both the monocular baseline and the stereo triangulation learning network outperform the prior state-of-the-arts in 3D object detection and localization on the challenging KITTI dataset.
\end{abstract}


\input{introduction}

\input{related_work}
\input{approach}

\input{metric_learning}
\input{implementation}

\input{experiment}
\vspace{-0.6cm}
\section{Conclusion}
We present a novel network for performing accurate 3D object detection using stereo information. We build a solid baseline monocular detector, which is flexibly extended to stereo by combining with the proposed TLNet. The key idea is to use 3D anchors to construct geometric correspondences between its projections in stereo images, from which the network learns to triangulate the targeted object in a forward pass. We also introduce an efficient channel reweighting method to strengthen informative features and weaken the noisy signals. All of these are integrated into our baseline detector and achieve state-of-the-art performance.

{\small  
\bibliographystyle{ieee}
\bibliography{reference_cvpr19}
} 

\end{document}

%% file: introduction.tex
\section{Introduction}
3D object detection aims at localizing amodal 3D bounding boxes of objects of specific classes in the 3D space~\cite{cai2013fastdetection, zeng2017multiview, lahond2017twoddriven, ren2016layout}. 
The detection task is relatively easier~\cite{engelcke2017vote3d, li20173dfcn, kim2018lidar} when active 3D scan data are provided. However, the active scan data are of high cost and limited in scalability.
We address the problem of 3D detection from passive imagery data, which require only low-cost hardware, adapt to different scales of objects, and offer fruitful semantic features.  

Monocular 3D detection with a single RGB image is highly ill-posed because of the ambiguous mapping from 2D images to 3D geometries, but still attracts research efforts~\cite{chen2016monocular,xu2018multifusion} because of its simplicity of design. Adding more input views can provide more information for 3D reasoning, which is ubiquitously demonstrated in the traditional multi-view geometry~\cite{hartley2003multiple} area by finding dense patch-level correspondences for points and then estimate their 3D locations by triangulation. The geometric methods deal with patch-level points with local features, while no semantic clues at object-level are considered.

Stereo data with pairs of images are more suitable for 3D detection, since the disparities between left and right images can reveal spacial variance, especially in the depth dimension. While extensive work~\cite{zbontar2016stereo, mayer2016dispnet, cheng2018depth} has been done on deep-learning based stereo matching, their main focus is on the pixel level rather than object level. 3DOP~\cite{chen20153dop} uses stereo images for 3D object detection and achieves state-of-the-art results. Nevertheless, later we show that we can obtain comparable results using \emph{only} a monocular image, by properly placing 3D anchors and extending the region-proposal network (RPN)~\cite{ren2017faster} to 3D. Therefore, we are motivated to reconsider how to better exploit the potential of stereo images for accurate 3D object detection.

\begin{figure}[t]
	\centering
	\includegraphics[width=1\linewidth]{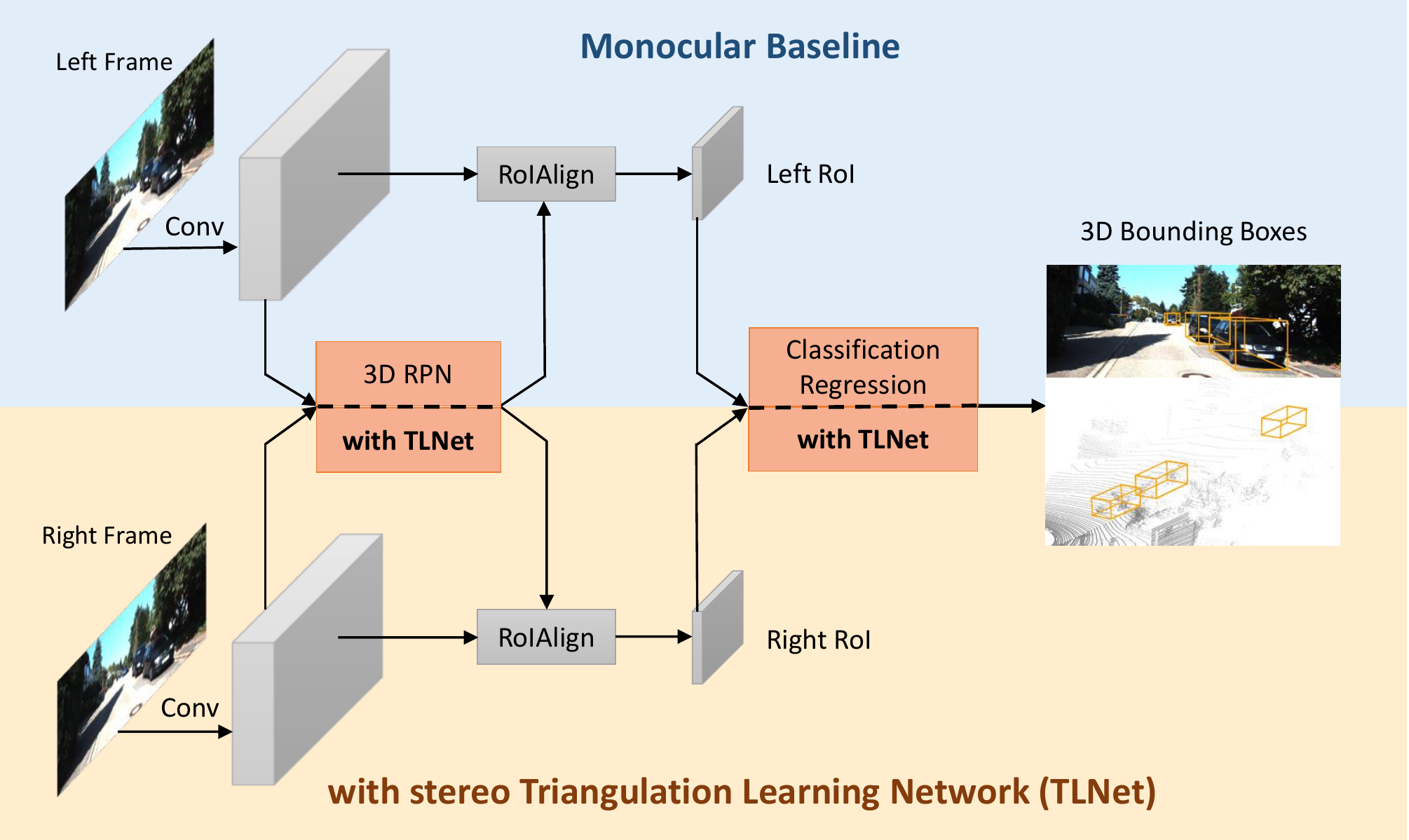}
	\label{fig:overview_simple}
	\caption{\textbf{Overview of the proposed 3D detection pipeline.} The baseline monocular network is indicated with blue background, and can be easily extended to stereo inputs by duplicating the baseline and further integrating with the proposed TLNet.}
	\end{figure}

In this paper, we propose the stereo \textbf{T}riangulation \textbf{L}earning \textbf{N}etwork (TLNet) for 3D object detection from stereo images, which is free of computing pixel-level depth maps and can be easily integrated into the baseline monocular detector. The key idea is to use a 3D anchor box to explicitly construct object-level geometric correspondences of its two projections on a pair of stereo images, from which the network learns to triangulate a targeted object near the anchor. In the proposed TLNet, we introduce an efficient feature reweighting strategy that strengthens informative feature channels by measuring left-right coherence. The reweighting scheme filters out the signals from noisy and mismatched channels to facilitate the learning process, enabling our network to focus more on the key parts of an object. Without any parameters, the reweighting strategy imposes little on computational burden.

To examine our design, we first propose a solid baseline monocular 3D detector, with an overview shown in  \fig{\ref{fig:overview_simple}}. In combination with TLNet, we demonstrate that significant improvement can be achieved in 3D object detection and localization in various scenarios. 
Additionally, we provide quantitative analysis of the feature reweighting strategy in TLNet to have a better understanding of its effects. In summary, our contributions are three-fold:

\begin{itemize}
\item A solid baseline 3D detector that takes only a monocular image as input, which has comparable performance with its state-of-the-art stereo counterpart.

\item A triangulation learning network that leverages the geometric correlations of stereo images to localize targeted 3D objects, which outperforms the baseline model by a significant margin on the challenging KITTI~\cite{geiger2012kitti} dataset.

\item A feature reweighting strategy that enhances informative channels of view-specific RoI features, which benefits triangulation learning by biasing the network attention towards the key parts of an object.
\end{itemize}





%% file: related_work.tex
\section{Related Work}
\paragraph{Monocular 3D Detection.}
Due to the information loss in the depth dimension, 3D object detection is in particular difficult given only a monocular image. 
Mono3D \cite{chen2016monocular} integrates semantic segmentation and context priors to generate 3D proposals. Extra computation is needed for semantic and instance segmentation, which slows down its running time.
Xu \etal \cite{xu2018multifusion} utilizes a stand-alone disparity estimation model~\cite{mahjourian2018unsupervised} to generate 3D point clouds from a monocular image, and then perform 3D detection using multi-level concatenated RoI features from RGB and the point cloud maps. However, it has to use extra data, i.e., ground truth disparities, to train the stand-alone model. Other methods~\cite{kehl2017ssd6d} leverage 3D CAD models to synthesize 3D object templates to supervise the training and guide the geometric reasoning in inference. While the previous methods are dependent on additional data for 3D perception, the proposed monocular baseline model can be trained using only ground truth 3D bounding boxes, which saves considerations on data acquisition. 

\paragraph{Multi-View based 3D Detection.}
MV3D~\cite{chen2017multiview} takes RGB images and multi-view projections of LIDAR 3D points as input, and then fuse the RoI pooled features for 3D bounding box prediction. By projecting the point clouds to bird's-eye-view (BEV) maps, it first generates 2D bounding box proposals in BEV, then fixes the height to obtain 3D proposals. RoIAlign~\cite{he2017mrcn} is performed in front-view and BEV point cloud maps and also in image to extract multi-view features, which are used to predict object classes and regress the final 3D bounding box. AVOD~\cite{ku2017joint} aggregates features from front view and BEV of LIDAR point to jointly generate proposals and detect objects. It is an early-fusion method, where the multi-view features are fused before the region-proposal stage \cite{he2017mrcn} to increase the recall rate. The most related approach to ours is 3DOP~\cite{chen20153dop} that uses stereo images for object detection. However, 3DOP~\cite{chen20153dop} directly relies on the disparity maps calculated from image pairs, resulting in its high computational cost and the imprecise estimation at distant regions. Our network is free of calculating pixel-level disparity maps. Instead, it learns to triangulate the target from left-right RoIs.

\paragraph{Learning based Stereo.}
Zbontar and LeCun~\cite{zbontar2016stereo} propose a stereo matching network to learning a similarity measure on small image patches to estimate the disparity at each location of the input image. Because comparing the image patches for each disparity would increase the computational cost combinationally, the authors propose to propagate the full-resolution images to compute the matching cost for all pixels in only one forward pass. DispNet~\cite{mayer2016dispnet} concatenates the stereo image pairs as input to learn the disparity and scene flow. Chen \etal \cite{cheng2018depth} design a convolutional spatial propagation network to estimate the affinity matrix for depth estimation. While the existing studies mainly focus on pixel-level learning, we primarily explore the instance-level learning for 3D object detection using the intrinsic geometric correspondence of the stereo images.

\paragraph{Triangulation.}
Triangulation localizes a point by forming triangles to it from known points, which is of universal applications in 3D geometry estimation~\cite{dai2018triangulation, petroff2018triangulation, herrera2014slam, nir2004causal}. In RGB-D based SLAM~\cite{dai2018triangulation}, triangulation can be utilized to create a sparse graph modeling the correlations of points and retrieve accurate navigation information in complex environment. It is also useful in camera calibration and motion tracking~\cite{herrera2014slam}. In stereo vision, typical triangulation~\cite{zbontar2016stereo} requires matching the 2D features in left and right frames at pixel level to estimate spacial dimensions, which can be computationally expensive and time consuming~\cite{mayer2016dispnet}. To avoid such computation and fully exploit stereo information for 3D object detection, we propose using 3D anchors as reference to detect and localize the object via learnable triangulation in a forward pass.

%% file: approach.tex
\section{Approach}
\begin{figure}
	\centering
	\includegraphics[width=1\linewidth]{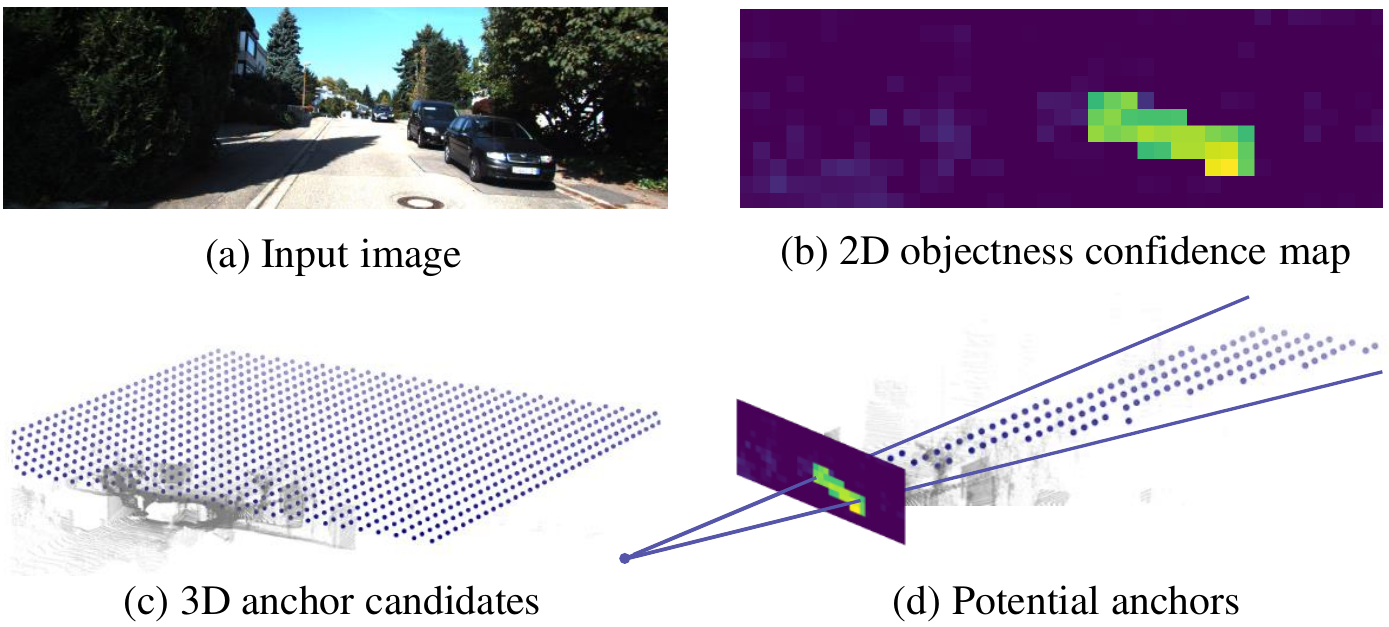}
	\caption{\textbf{Front view anchor generation.} Potential anchors are of high objectness in the front view. Only the potential anchors are fed into RPN to reduce searching space and save computational cost.}
	\label{fig:fvanchors}
\end{figure}

In this section, we first present the baseline model that predicts oriented 3D bounding boxes from a base image $I$, and then introduce the triangulation learning network integrated in the baseline for a pair of stereo image $(I^l, I^r)$.
An oriented 3D bounding box is defined as $B=(\mathbf{C}, \mathbf{S}, \theta)$, where $\mathbf{C}=(c_x, c_y, c_z)$ is the 3D center, $\mathbf{S}=(h, w, l)$ is the size along axis, and $\theta$ is the angle between itself and the vertical axis.

\subsection{Baseline Monocular Network}
The baseline network taking a monocular image as the input is composed of a backbone and three subsequent modules, \ie, the front view anchor generation, the 3D box proposal and refinement. The three-stage pipeline progressively reduces the searching space by selecting confident anchors, which highly reduces computational complexity.

\subsubsection{Front View Anchor Generation}
\label{sssec:frontviewanchors}
Following the dimension reduction principle, we first reduce the searching space in the 2D front view. 
The input image $I$ is divided into a $G_x \times G_y$ grid, where each cell predicts its objectness. 
The output objectness represents the confidence how likely the cell is surrounded by 2D projections of targeted objects. An example of the confidence map is shown in \fig{\ref{fig:fvanchors}}. 
In training, we first calculate 2D projected centers of the ground truth 3D boxes and compute their least Euclidean distance to all cells in the $G_x \times G_y$ grid. Cells with distances less than 1.8 times their width are considered as foreground. For inference, foreground cells are selected to generate potential anchors. 

Thus, we obtain 3D anchors located at the rays issued from the potential cells in an anchor pool, which contains a set of 3D anchors uniformly sampled at an interval of 0.25m on the ground plane within the view frustum and depth ranging $\left[0,70\right]$ meters. Anchors are represented by its 3D center and the prior size along three axes. There are two anchors with BEV orientation $0^\circ$ and $90^\circ$ for each object class at the same location. For different classes, we calculate the prior size by averaging corresponding samples in the training split. An example of generated anchors are shown in \fig{\ref{fig:fvanchors}}.

\begin{figure}
	\centering
	\includegraphics[width=6cm]{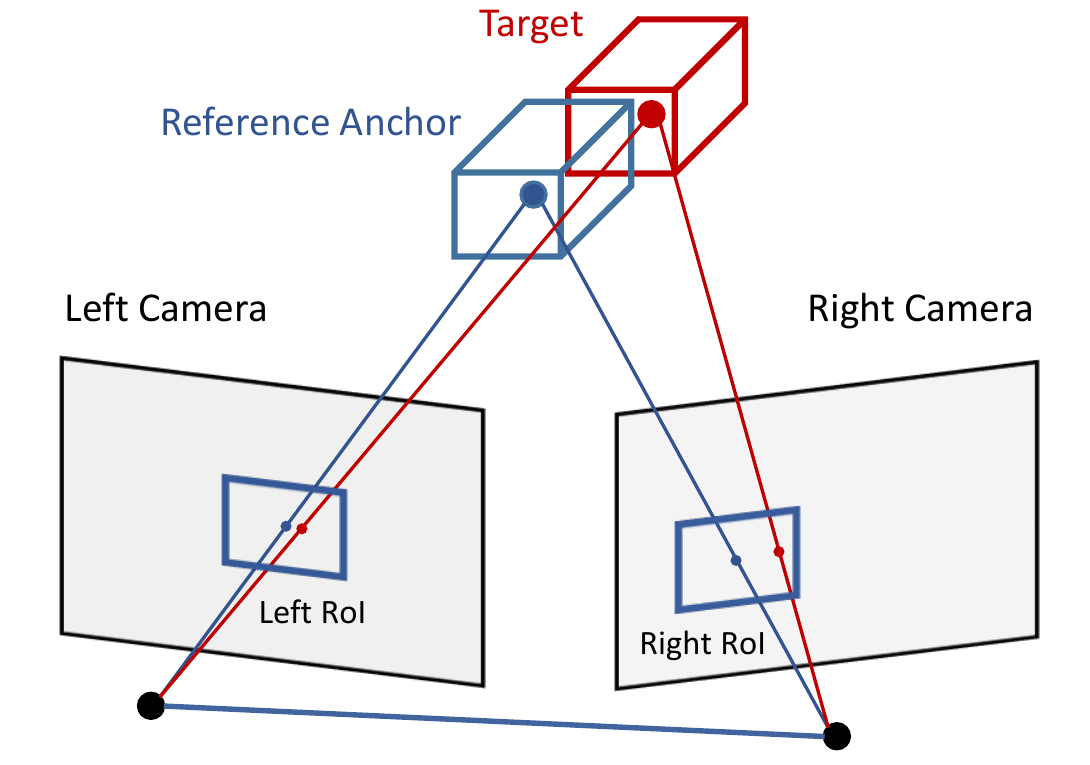}
	\caption{\textbf{Anchor triangulation.} By projecting the 3D anchor box to stereo images, we obtain a pair of RoIs. The left RoI establishes a geometric correspondence with the right one via the anchor box. The nearby target is present in both RoIs with slightly positional differences. Our TLNet takes the RoI pair as input and utilizes the 3D anchor as reference to localize the targeted object.}
	\label{fig:instanceleveldisparity}
\end{figure}

\subsubsection{3D Box Proposal and Refinement}
\label{sssec:proposalsandrefinement}
The multi-stage proposal and refinement mechanism in our baseline network is similar to Faster-RCNN~\cite{ren2017faster}.
In the 3D RPN, the potential anchors from front view generation are projected to the image plane to obtain RoIs. RoIAlign~\cite{he2017mrcn} is adopted to crop and resize the RoI features from the feature maps. Each crop of RoI features is fed to task specific fully connected layers to predict 3D objectness confidence as well as regress the location offsets $\Delta \mathbf{C} = (\Delta c_x, \Delta c_y, \Delta c_z)$ and dimension offsets $\Delta \mathbf{S} = (\Delta h, \Delta w, \Delta l)$ to the ground truth. Non-Maximum Supression (NMS) is adopted to keep the top $K$ proposals, where $K=1024$ for both training and inference. In the refinement stage, we project those top proposals to image and again use RoIAlign~\cite{he2017mrcn} to crop and resize the region of interest. The RoI features are passed to fully connected layers that classify the object and regress the 3D bounding box offsets $(\Delta \mathbf{C},\Delta \mathbf{S})$ as well as the orientation vector $\mathbf{v}_{\theta}$in local coordinate system as defined in \cite{ku2017joint}.

%% file: metric_learning.tex
\subsection{Triangulation Learning Network}
The stereo 3D detection is performed by integrating a triangulation learning network into the baseline model.
In the following, we first introduce the mechanism of the TLNet that focuses on object-level triangulation in opposite to computationally expensive pixel-level disparity estimation, and then present the details of its network architecture.

\begin{figure*}
	\centering
	\includegraphics[width=1\linewidth]{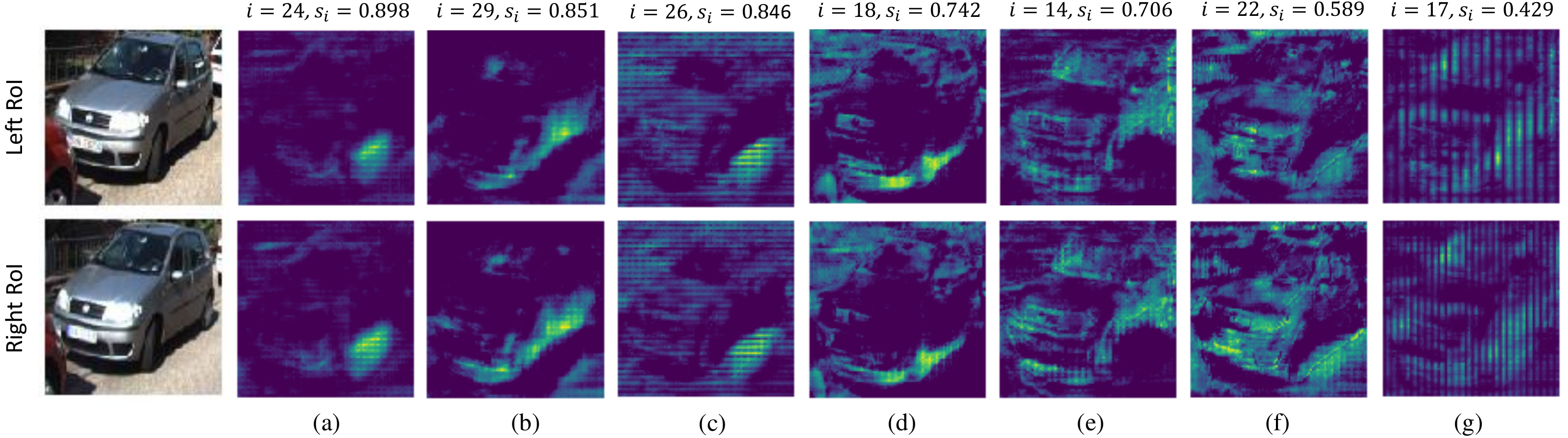}
	\caption{\textbf{Activations of different channels with different coherence scores.} These small feature maps are cropped out using RoIAlign from the last convolutional layer, where the first row is from the left branch and the second row is from the right. Two branches of convolutional layers share the weights. Coherence score $s_i$ is calculated for channel $i$. As we can see, channel 17 (g) and 22 (f) have noisy and less concentrated activations, while chanel 24 (a) is clear and informative of the key points of an object, \eg, wheels. Our objective is enhancing channels like (a) and weaken those like (g) and (f), so as to focus the network attention on specific parts of the object, which is empirically beneficial for discerning slight positional difference between where the object is present in left and right RoIs.}
	\label{fig:roipatches}
\end{figure*}

\begin{figure}
	\centering
	\includegraphics[width=1\linewidth]{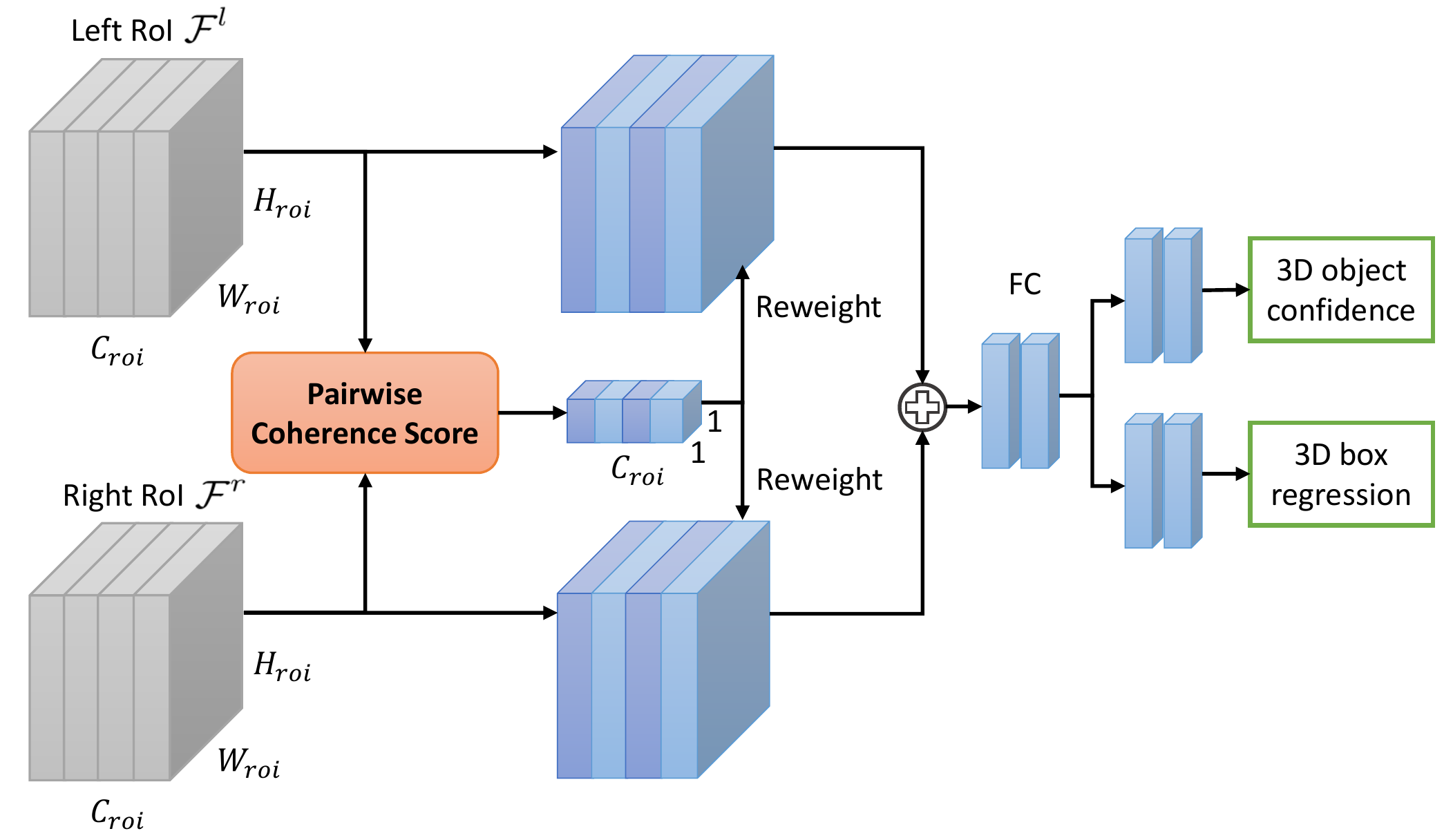}
	\caption{\textbf{TLNet architecture.} The inputs are a pair of RoIs obtained by projecting a 3D anchor box to the left and right feature maps with $C_{roi}$ channels. The coherence score $s_i$ is computed between each left channel $\mathcal{F}^{l}_i$ and right channel $\mathcal{F}^{r}_i$. We \emph{reweight} the $i\textsuperscript{th}$ channel by multiplying with $s_i$. The features are fused and fed to fully-connected layers to predict objectness confidence and 3D bounding box offsets to the reference anchor.}
	\label{fig:metriclearning}
\end{figure}

\subsubsection{Anchor Triangulation}
Triangulation is known as localizing 3D points from multi-view images in the classical geometry fields, while our objective is to localize a 3D object and estimates its size and orientation from stereo images. To achieve this, we introduce an \emph{anchor triangulation} scheme, in which the neural network uses 3D anchors as reference to triangulate the targets. 

Considering the RPN stage, we project the pre-defined 3D anchor to stereo images and obtain a pair of left-right RoIs, as illustrated in \fig{\ref{fig:instanceleveldisparity}}. If the anchor tightly fits the target in 3D, its left and right projections can consistently bound the object in 2D. On the contrary, when the anchor fails in fitting the the object, their geometric differences in 3D are reflected in the visual disparity of the left-right RoI pair. 
The 3D anchor explicitly constructs correspondences between its projections in multiple views. 
Since the location and size of the anchor box are already known, modeling the anchor triangulation is conducive to estimating the 3D objectness confidence, \ie, how well the anchor matches a target in 3D, as well as regressing the offsets applied to the box to minimize its variance with the target. Therefore, we propose TLNet towards triangulation learning as follows.

\subsubsection{TLNet Architecture}
\label{subsubsec:tlnet}
The TLNet takes as input a pair of left-right RoI features  $\mathcal{F}^{l}$ and $\mathcal{F}^{r}$ with $C_{roi}$ channels and size $H_{roi} \times W_{roi}$, which are obtained using RoIAlign \cite{he2017mrcn} by projecting the same 3D anchor to the left and right frames, as shown in \fig{\ref{fig:metriclearning}}. We utilize the left-right \emph{coherence scores} to \emph{reweight} each channel. The reweighted features are fused using element-wise addition and passed to task-specific fully-connected layers to predict the objectness confidence and 3D bounding box offsets, i.e., the 3D geometric variance between the anchor and target.

\paragraph{Coherence Score.} The positional difference between where the target is present in the left and right RoIs reveals the spacial variance between the target box and the anchor box, as illustrated in \fig{\ref{fig:instanceleveldisparity}}. The TLNet is expected to utilize such a difference to predict the relative location of the target to the anchor, estimate whether they are a good match, \ie, objectiveness confidence, and regress the 3D bounding box offsets. To enhance the discriminative power of spacial variance, it is necessary to focus on representational key points of an object. \fig{\ref{fig:roipatches}} shows that, though without explicit supervision, some channels in the feature maps has learned to extract such key points, \eg, wheels.
The coherence score $s_i$ for the $i^{th}$ channel is defined as:
\begin{align}
&s_i =  cos<\mathcal{F}^{l}_i, \mathcal{F}^{r}_i> = \frac{\mathcal{F}^{l}_i \cdot \mathcal{F}^{r}_i} {||\mathcal{F}^{l}_i||  \cdot ||\mathcal{F}^{r}_i||}
\label{eq:simi}
\end{align}
where $cos$ is the cosine similarity function for each channel, $\mathcal{F}^{l}_i$ and $\mathcal{F}^{r}_i$ are the $i^{th}$ pair of feature maps, \ie, features from the $i^{th}$ channel in left and right RoIs. 

From \fig{\ref{fig:roipatches}}, we observe that the coherence score $s_i$ is lower when the activations are noisy and mismatched, while it is higher if the left-right activations are clear and coherent. In fact, from a mathematical perspective, $\mathcal{F}^{l}_i$ and $\mathcal{F}^{r}_i$ can be viewed as a pair of signal vectors. As you flatten along the row dimension, consecutive activations are more likely to align with other consecutive activations in a pair of RoIs, \ie, $s_i$ is closer to $1$.



\paragraph{Channel Reweighting.} By multiplying the $i^{th}$ channel with $s_i$, we weaken the signals from noisy channels and bias the attention of the following fully-connected layers towards coherent feature representations. The reweighting is done in pairs for both the left and right feature maps, taking the form:
\begin{align}
&\mathcal{F}^{l, re}_i = s_i \mathcal{F}^{l}_i, \quad \mathcal{F}^{r, re}_i = s_i \mathcal{F}^{r}_i
\label{eq:reweight}
\end{align}
, and it is implemented in the TLNet as illustrated in \fig{\ref{fig:metriclearning}}. We will demonstrate that the reweighting strategy has a positive effect on triangulation learning in the experiments.



\paragraph{Network Generality.} The proposed architecture can be easily integrated into the baseline network by replacing the fully-connected layers after RoIAlign, in both the RPN and the refinement stage. In the refinement stage, classification outputs are object classes instead of objectness confidence as is in RPN. For the CNN backbone, the left and right branches share their parameters. Computing cosine similarity and reweighting do not require any extra parameters, thus imposing little in memory burden. SENet \cite{hu2018senet} also involves reweighting feature channels. Their goal is to model the cross-channel relationships of a single input, while ours is to measure the coherence of a pair of stereo inputs and select confident channels for triangulation learning. In addition, their weights are learned by fully-connected layers, whose behavior is less interpretable. Our weights are the pairwise cosine similarities with clear physical significance. $\mathcal{F}^{l}_i$ and $\mathcal{F}^{r}_i$ can be viewed as two vectors, and $s_i$ describes their included angle, i.e., correlation.


%% file: implementation.tex
\subsection{Implementation Details}
\paragraph{Network Setup.}
We choose VGG-16 \cite{matthew2014vgg} as the CNN backbone, but without its fully-connected layers after $pool5$. Parameters are shared in the left and right branches. All the input images are resized to $384 \times 1248$ so that they can be divided by 2 for at least 5 times. For the front-view objectness prediction, we use the left branch only. We apply $1 \times 1$ convolution to the output feature maps of $pool5$ and reduce the channels to 2, indicating the background and objectness confidence. Each pixel in the output feature maps represents a grid cell and yields a prediction.

For the region proposal and refinement stage, we start from the output of $conv4$ rather than $pool5$. To improve the performance on small objects, we leverage Feature Pyramid \cite{lin2016fpn} to upsample the feature maps to the original resolution. Since the region proposal stage aims at filtering background anchors and keep the confident anchors in an efficient manner, channels of the feature maps are reduced to 4 using $1 \times 1$ convolution before RoIAlign \cite{he2017mrcn} and fully connected layers to save computational cost. For the refinement stage, however, we use full channels containing more information to yield finer predictions. 

\paragraph{Training.}
All the weights are initialized by Xavier initializer \cite{xavier2010init} and no pretrained weights are used. L2 regularization is applied to the model parameters with a decay rate of 5e-3. We first train the front-view objectness map for 20K iterations, then along with the RPN for another 40K iterations. In the next 60K iterations we add the refinement stage. For the above 120K iterations, the network is trained using Adam optimizer \cite{kingma:2014} at a learning rate of 1e-4. Finally, we use SGD to further optimize the network for 20K iterations at the same learning rate. The batchsize is always set to 1 for the input. The network is trained using a single GPU of NVidia Tesla P40.

%% file: experiment.tex
\begin{table*}[!htbp]   
	\centering
	\setlength{\tabcolsep}{3mm}{  
	\begin{spacing}{1.1}
			\scalebox{0.75}{
				\begin{tabular}{|p{3cm}<{\centering}|p{1cm}<{\centering}||ccc||ccc||ccc|}
					\hline
					\multirow{2}{*}{Method} & \multirow{2}{*}{Data}           & \multicolumn{3}{c||}{AP\textsubscript{3D}(IoU=0.3)}                                 & \multicolumn{3}{c||}{AP\textsubscript{3D}(IoU=0.5)}                             & \multicolumn{3}{c|}{AP\textsubscript{3D}(IoU=0.7)}                  \\ \cline{3-11} 
				                                         	&                                               & \multicolumn{1}{c|}{Easy}     & \multicolumn{1}{c|}{Moderate}  & Hard               & \multicolumn{1}{c|}{Easy} & \multicolumn{1}{c|}{Moderate} & Hard                & \multicolumn{1}{c|}{Easy} & \multicolumn{1}{c|}{Moderate} & Hard           \\ \hline \hline
                      
					VeloFCN                 & LiDAR                                                   & /                             & /                              & /                      & 67.92                          & 57.57              & 52.56                     & 15.20                         & 13.66               & 15.98 \\
					\hline
					Mono3D                  & Mono                                                & 28.29                          &23.21                           &19.49               & 25.19                     & 18.20                         & 15.22               & 2.53                      & 2.31                          & 2.31           \\ 
					MF3D                    & Mono                                                &/                              &/                               &/                   & 47.88                     & 29.48                         & 26.44               & {10.53}                   & 5.69                          & 5.39           \\ 
					MonoGRNet               & Mono                                                & 72.17 &59.57 &46.08 &50.51 &36.97 &30.82 &13.88 &10.19 &7.62 \\
                    3DOP                    & Stereo                                                & 69.79                         &52.22                           &49.64               & 46.04                     & 34.63                         & 30.09               & 6.55                      & 5.07                          & 4.10           \\   \hline       
                    Ours (baseline)         & Mono                                            & 72.91                         & 55.72                          & 49.19              &48.34                      & 33.98                         & 28.67               & 13.77                     & 9.72                          & 9.29 \\ 
                
                    Ours                    & Stereo                                           & \textbf{78.26}                & \textbf{63.36}                 & \textbf{57.10}     &\textbf{59.51}             & \textbf{43.71}                & \textbf{37.99}      & \textbf{18.15}            & \textbf{14.26}                & \textbf{13.72} \\ \hline
				\end{tabular}  
			} 
			\end{spacing}
     }
	\caption{\textbf{3D detection performance.} Average Precision of 3D bounding boxes on KITTI~\cite{geiger2012kitti} validation set. The LiDAR based method VeloFCN~\cite{li2016} is listed for reference but not compared.
	}
	\label{tab:3dap} 
\end{table*}

\begin{table*}[!htbp]   
	\centering
	\setlength{\tabcolsep}{3mm}{  
	\begin{spacing}{1.1}
			\scalebox{0.75}{
				\begin{tabular}{|p{3cm}<{\centering}|p{1cm}<{\centering} ||ccc||ccc||ccc|}
					\hline
					\multirow{2}{*}{Method} & \multirow{2}{*}{Data}       & \multicolumn{3}{c||}{AP\textsubscript{BEV}(IoU=0.3)}                                 & \multicolumn{3}{c||}{AP\textsubscript{BEV}(IoU=0.5)}                             & \multicolumn{3}{c|}{AP\textsubscript{BEV}(IoU=0.7)}                  \\ \cline{3-11} 
				                                           &                                             & \multicolumn{1}{c|}{Easy}     & \multicolumn{1}{c|}{Moderate}  & Hard               & \multicolumn{1}{c|}{Easy} & \multicolumn{1}{c|}{Moderate} & Hard                & \multicolumn{1}{c|}{Easy} & \multicolumn{1}{c|}{Moderate} & Hard           \\ \hline \hline
                    VeloFCN                 & LiDAR                                   &  /                            & /                               & /                  & 79.68                    & 63.82                         & 62.80               & 40.14                     & 32.08                         & 30.47          \\
				    \hline
					Mono3D                  & Mono                                    & 32.76                         &25.15                           &23.65               & 30.50                     & 22.39                         & 19.16               & 5.22                      & 5.19                          & 4.13           \\ 
					MF3D                    & Mono                                     & /                             & /                              & /                  & 55.02                     & 36.73                         & 31.27               & 22.03                     & 13.63                         & 11.60           \\ 
					MonoGRNet               & Mono                                                &73.10 &60.66 &46.86 &54.21 &39.69 &33.06 &24.97 &19.44 &16.30 \\
                    3DOP                    & Stereo                                       & 71.41                         &57.78                           &51.91               & 55.04                     & 41.25                         & 34.55               & 12.63                     & 9.49                          & 7.59           \\          
                    \hline
                    Ours (baseline)         & Mono                                   & 74.18                         & 57.04                          & 50.17              & 52.72                     & 37.22                         & 32.16               & 21.91                     & 15.72                          & 14.32 \\
                   
                    Ours                    & Stereo                                  & \textbf{81.11}                & \textbf{65.25}                 & \textbf{58.15}     & \textbf{62.46}            & \textbf{45.99}                & \textbf{41.92}      & \textbf{29.22}            & \textbf{21.88}                 & \textbf{18.83} \\ \hline
				\end{tabular}  
			} 
			\end{spacing}
     }
	\caption{\textbf{BEV detection performance.} Average Precision of BEV bounding boxes on KITTI~\cite{geiger2012kitti} validation set. Different from typical 2D detection evaluation, the bounding boxes are orientated, \ie, not necessarily aligned on each axis.
	}
	\label{tab:bvap} 
\end{table*}

\begin{table*}[!htbp]   
	\centering
	\setlength{\tabcolsep}{3mm}{  
	\begin{spacing}{1.1}
			\scalebox{0.75}{
				\begin{tabular}{|p{3cm}<{\centering}|p{1cm}<{\centering} ||ccc||ccc||ccc|}
					\hline
					\multirow{2}{*}{Method} & \multirow{2}{*}{Data}       & \multicolumn{3}{c||}{AP\textsubscript{LOC}(\textless2m)}                                 & \multicolumn{3}{c||}{AP\textsubscript{LOC}(\textless1m)}                             & \multicolumn{3}{c|}{AP\textsubscript{LOC}(\textless0.5m)}                  \\ \cline{3-11} 
					                              &                                                        & \multicolumn{1}{c|}{Easy}     & \multicolumn{1}{c|}{Moderate}  & Hard               & \multicolumn{1}{c|}{Easy} & \multicolumn{1}{c|}{Moderate} & Hard                & \multicolumn{1}{c|}{Easy} & \multicolumn{1}{c|}{Moderate} & Hard           \\ \hline \hline
                      
					Mono3D                  & Mono                                    & 47.21                         &35.82                           &35.40                       & 17.74                     & 14.80                         & 14.05               & 4.78                      & 4.06                          & 4.03           \\ 
					MonoGRNet               & Mono                                                &83.09 &64.82 &55.79 &56.29 &42.29 &35.01 &28.38 &19.73 &18.06 \\
                    3DOP                    & Stereo                                       & 84.53                         &65.37                           &64.10                     & 60.84                     & 45.11                         & 40.35               & 27.53                     & 19.39                          & 17.64           \\          
                    \hline
                    Ours (baseline)         & Mono                                   & 75.50                         & 59.33                          & 52.92                      & 58.11                     & 41.05                         & 36.59               & 30.82                     & 21.25                          & 18.03 \\
                    Ours                        & Stereo                                  & \textbf{86.78}           & \textbf{72.17}              & \textbf{66.27}        & \textbf{69.15}         & \textbf{51.68}            & \textbf{47.68}  & \textbf{37.36}        & \textbf{27.42}             & \textbf{24.56} \\ \hline
				\end{tabular}  
			} 
			\end{spacing}
     }
	\caption{\textbf{3D localization performance.} Average Location Precision of 3D bounding boxes under different distance thresholds on KITTI~\cite{geiger2012kitti} validation set.}
	\label{tab:3dloc} 
\end{table*}

\begin{figure*}[ht] 
	\centering
	\scriptsize
	\includegraphics[width=0.98\linewidth]{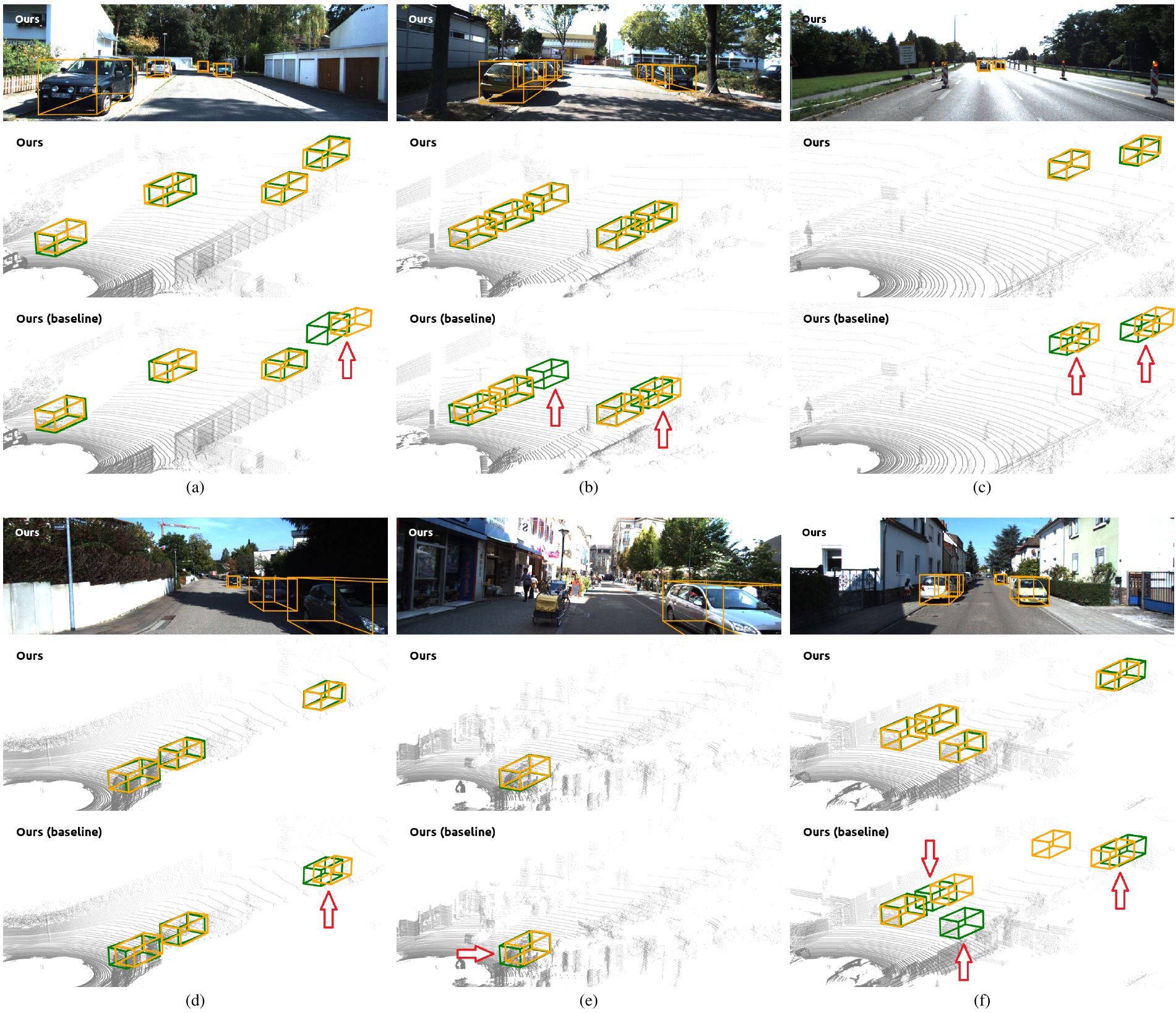}

\caption{\textbf{Qualitative comparison.} Orange bounding boxes are detection results, while the green boxes are ground truths. For our main method, we also visualize the projected 3D bounding boxes in image, i.e., the first and forth rows. The lidar point clouds are visualized for reference but not used in both training and evaluation. It is shown that the triangulation learning method can reduce missed detections and improve the performance of depth prediction at distant regions.}
	\label{fig:visualize3d}
\end{figure*}

\section{Experiment}
We evaluate the proposed network on the challenging KITTI dataset~\cite{geiger2012kitti}, which contains 7481 training images and 7518 testing images with calibrated camera parameters. Detection is evaluated in three regimes: easy, moderate and hard, according to the occlusion and truncation levels. We use the train1/val1 split setup in the previous works~\cite{chen2016monocular,chen2017multiview}, where each set contains half of the images. Objects of class $Car$ are chosen for evaluation. Because the number of cars exceeds that of other objects by a significant margin, they are more suitable to assess a data-hungry deep-learning network. We compare our baseline network with state-of-the-art monocular 3D detectors, MF3D~\cite{xu2018multifusion},  Mono3D~\cite{chen2016monocular} and MonoGRNet~\cite{qin2019monogr}. For stereo input, we compare our network with 3DOP~\cite{chen20153dop}. 

\paragraph{Metrics.}
For 3D detection, we follow the official settings of KITTI benchmark to evaluate the 3D Average Precision (AP\textsubscript{3D}) at different 3D IoU thresholds. To evaluate the bird's eye view (BEV) detection performance, we use the BEV Average Precision (AP\textsubscript{BEV}). We also provide results for Average Location Precision (AP\textsubscript{LOC}), in which a ground truth object is recalled if there is a predicted 3D location within certain distance threshold. Note that we obtain the detection results of Mono3D and 3DOP from the authors so as to evaluate them with different IoU thresholds. The evaluation of MF3D follows their original paper.

\vspace{-0.2cm}
\subsection{3D Object Detection}

\paragraph{Monocular Baseline.} 3D detection results are shown in \tab{\ref{tab:3dap}}. Our baseline network outperforms monocular methods MF3D~\cite{xu2018multifusion} and Mono3D~\cite{chen2016monocular} in 3D detection. For $moderate$ scenarios, our AP\textsubscript{3D} exceeds MF3D~\cite{xu2018multifusion} by $4.50\%$ at IoU = 0.5 and $4.65\%$ at IoU = 0.7. For $easy$ scenarios the gap is smaller. Since the raw detection results of MF3D~\cite{xu2018multifusion} are not publicly available, detailed quantitative comparison cannot be made. But it is possible that a part of the performance gap comes from the object proposal stage, where MF3D~\cite{xu2018multifusion} proposes on the image plane, while we directly propose in 3D. In $moderate$ and $hard$ cases where objects are occluded and truncated, 2D proposals can have difficulties retrieving the 3D information of a target that is partly unseen. In addition, the wide margin with Mono3D~\cite{chen2016monocular} reveals a better expressiveness of learnt features than handcrafted features.
\paragraph{TLNet.} By combining with the proposed TLNet, our method outperforms the baseline network and 3DOP~\cite{chen20153dop} across all 3D IoU thresholds in $easy$, $moderate$ and $hard$ scenarios. Under IoU thresholds of 0.3 and 0.5, stereo triangulation learning brings $\sim10\%$ improvement in AP\textsubscript{3D}. In comparison with 3DOP~\cite{chen20153dop}, our method achieves $\sim10\%$ better AP\textsubscript{3D} across all regimes. Note that 3DOP~\cite{chen20153dop} needs to estimate the pixel-level disparity maps from stereo images to calculate depths and generate proposals, which is erroneous at distant regions. Instead, the proposed method utilizes 3D anchors to construct geometric correspondence between left-right RoI pairs, then triangulates the targeted objects using coherent features. According to the curves in \fig{\ref{fig:apcurves}} (a), our main method has a higher precision than 3DOP~\cite{chen20153dop} and a higher recall than the baseline.

\subsection{BEV Object Detection}

\paragraph{Monocular Baseline.} Results are presented in \tab{\ref{tab:bvap}}. Compared with 3D detection, we still evaluate the same set of 3D bounding boxes, but the vertical axis is disregarded for BEV IoU calculation. The baseline keeps its advantages over MF3D in $moderate$ and $hard$ scenarios. Note that MF3D uses extra data to train pixel-level depth maps. In $easy$ cases, objects are clearly presented, and pixel-level predictions with sufficient local features can obtain more accurate results in statistics. However, pixel-level depth maps with high resolution are unfortunately not always available, indicating their limitations in real applications.

\vspace{-0.2cm}
\paragraph{TLNet.} Not surprisingly, by use of TLNet, we outperform the monocular baseline under all IoU thresholds in various scenarios. At IoU threshold 0.3 and 0.5, the triangulation learning yields $\sim8\%$ increase in AP\textsubscript{BEV}. Our method also surpasses 3DOP by a notable margin, especially for strict evaluation criteria, \eg, under IoU threshold of 0.7, which reveals the high precision of our predictions. Such performance improvement mainly comes from two aspects: 1) the solid monocular baseline already achieves comparable results with 3DOP; 2) the stereo triangulation learning scheme further enhances the capability of our baseline model in object localization. \fig{\ref{fig:apcurves}} (b) further compares the Recall-Precision curves under IoU threshold of 0.3.

\subsection{3D Localization}
\paragraph{Monocular Baseline.} Results are shown in \tab{\ref{tab:3dloc}}. Our monocular baseline achieve better results than 3DOP~\cite{chen20153dop} under the strict distance threshold of 0.5m, indicating the high precision of our top predictions. 

\begin{figure}[!htbp] 
	\centering
	\scriptsize
	\begin{tabular}{c@{\hspace{0.1cm}}c@{\hspace{0.1cm}}c}
		\includegraphics[width=0.33\linewidth]{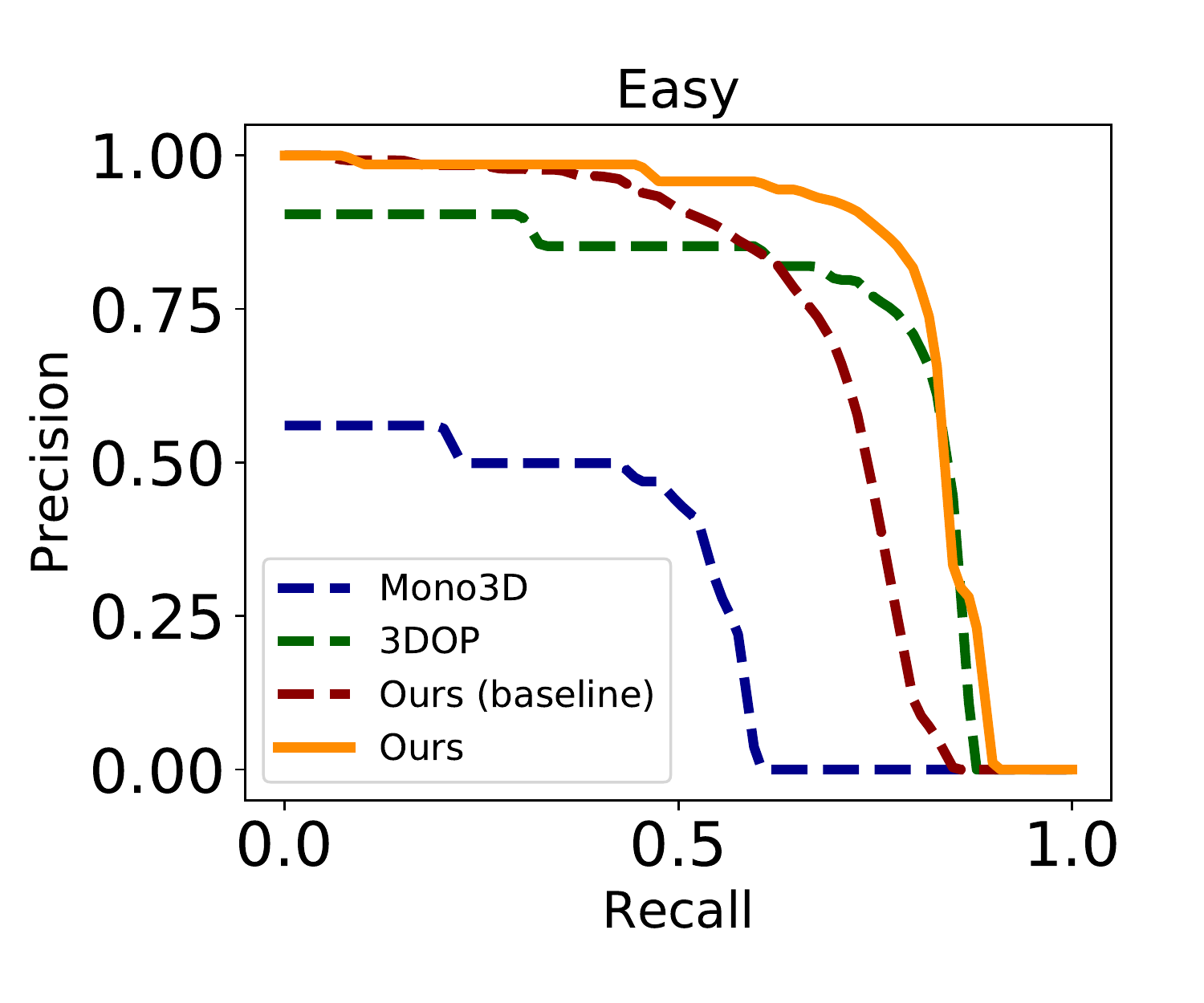} &
		\includegraphics[width=0.33\linewidth]{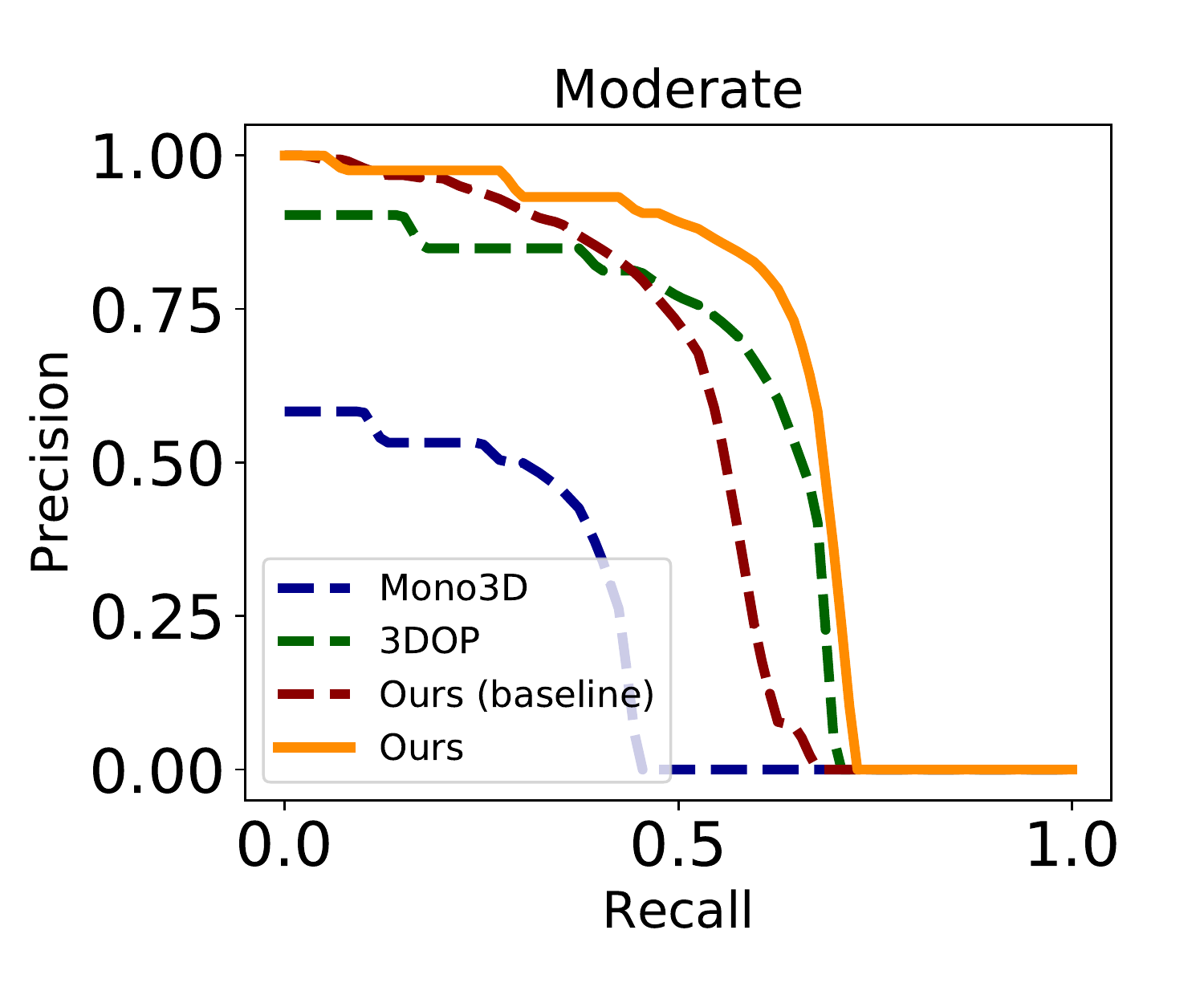} &
		\includegraphics[width=0.33\linewidth]{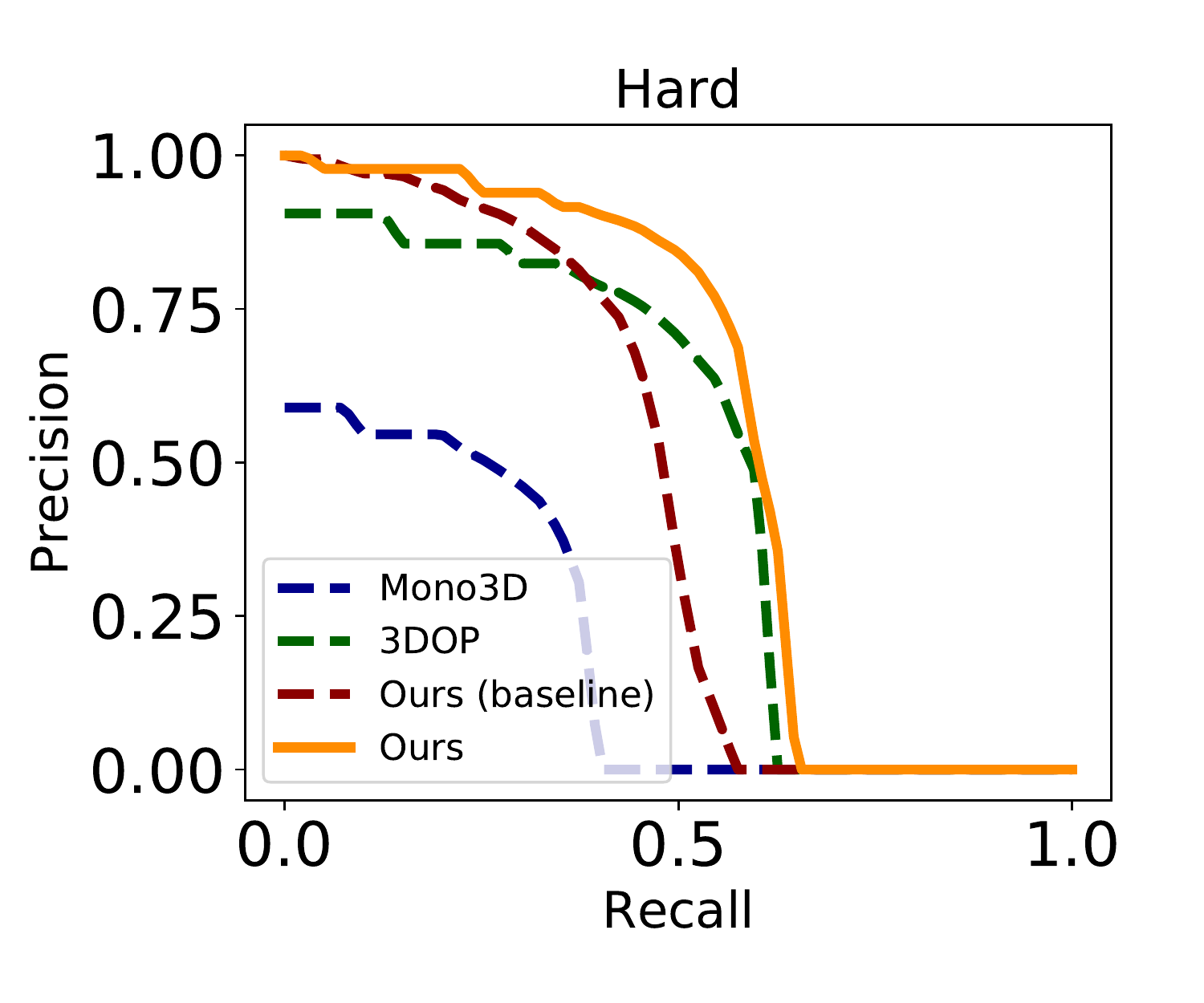} \\
		\multicolumn{3}{c}{(a) 3D object detection at IoU threshold of 0.3. }  \\
		
		\includegraphics[width=0.33\linewidth]{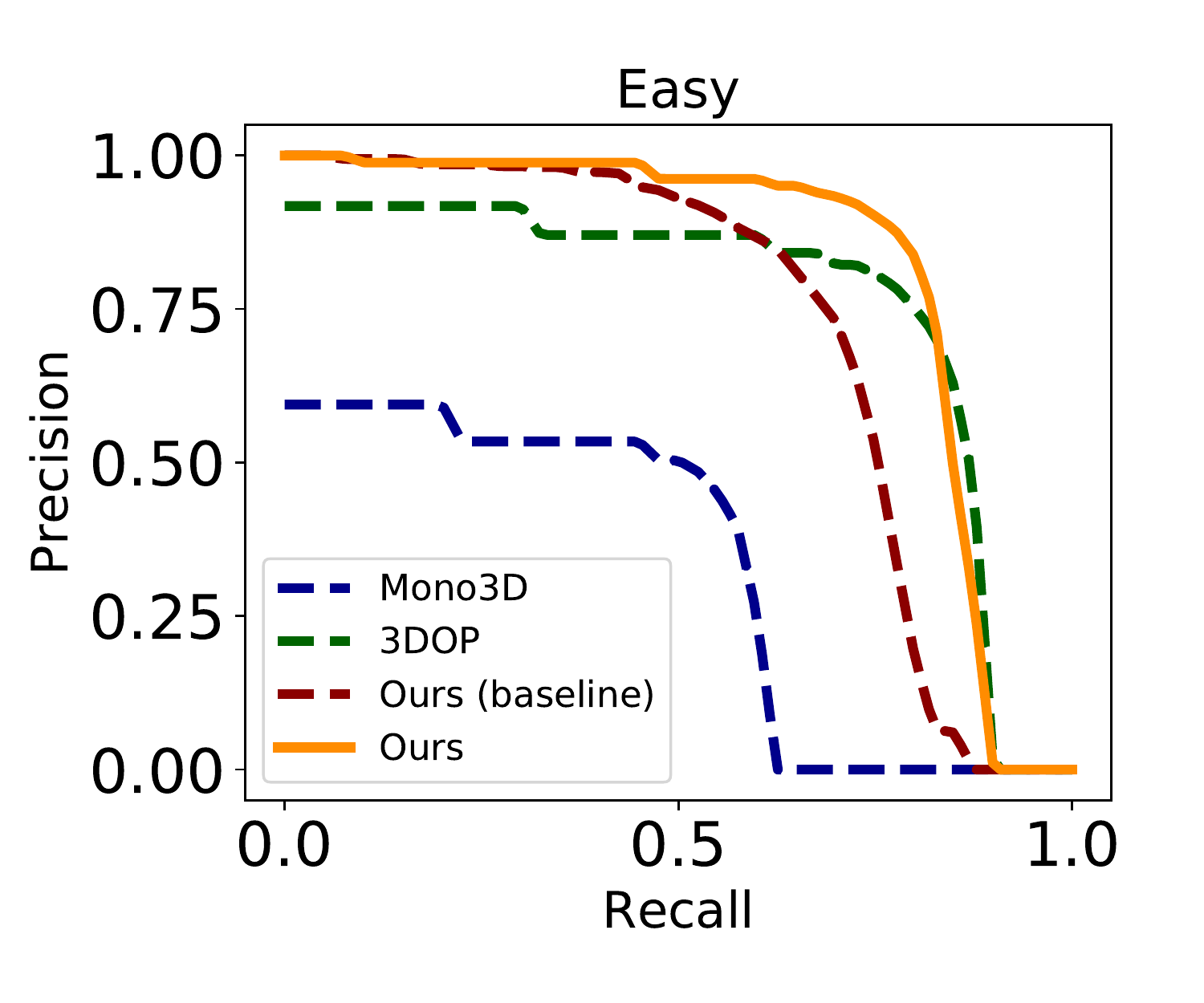} &
		\includegraphics[width=0.33\linewidth]{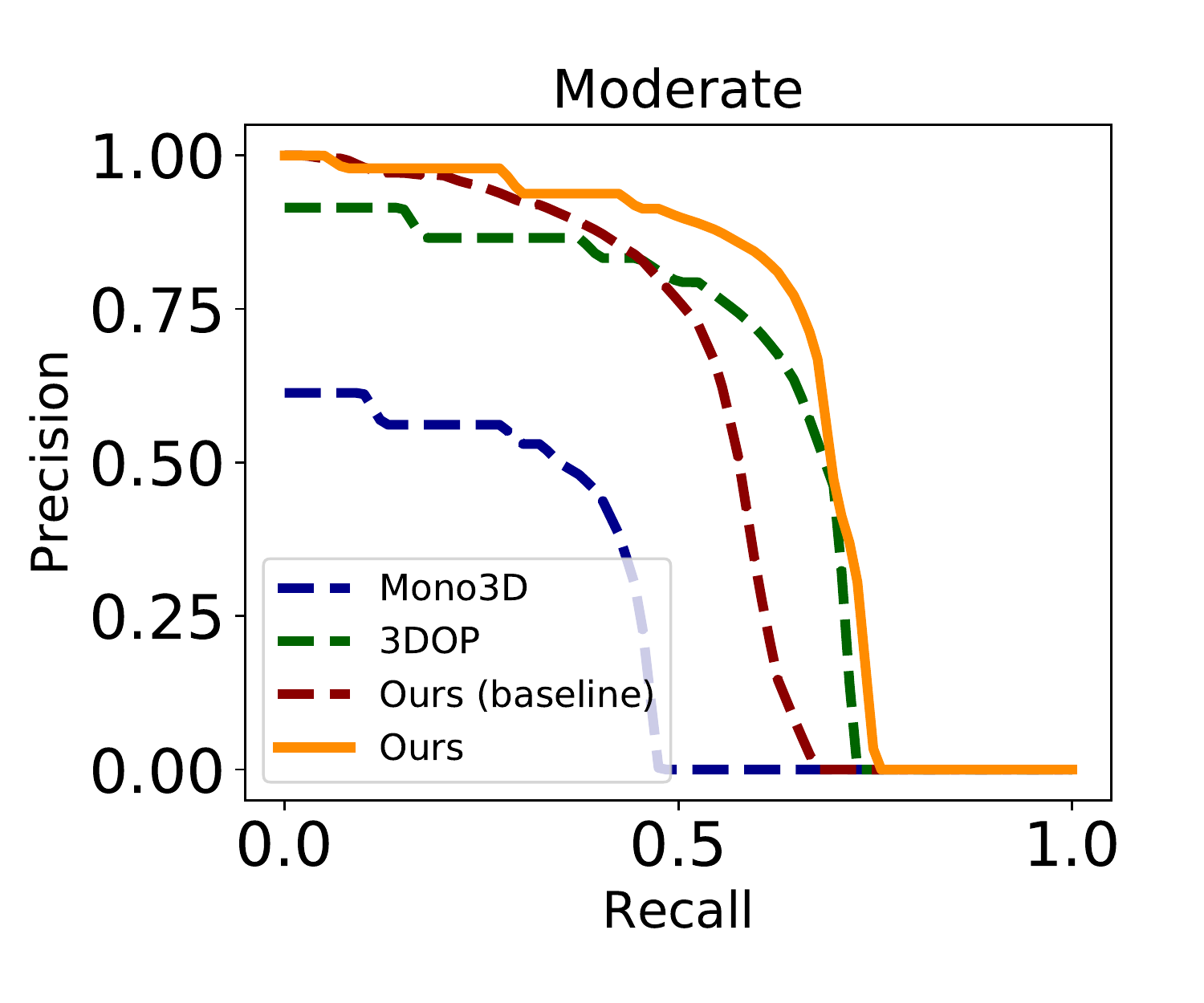} &
		\includegraphics[width=0.33\linewidth]{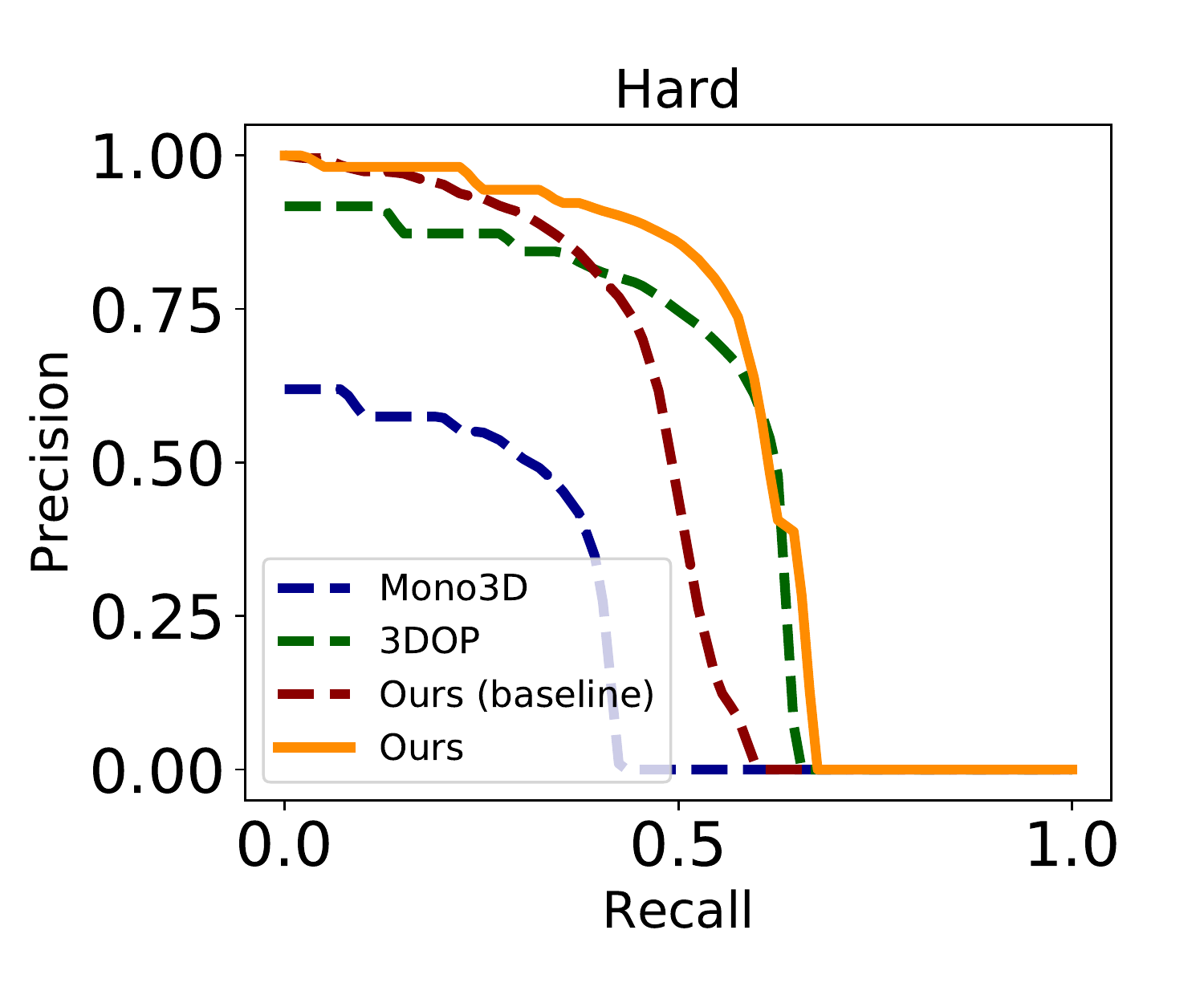} \\
		\multicolumn{3}{c}{(b) BEV object detection at IoU threshold of 0.3. }  \\
		
		\includegraphics[width=0.33\linewidth]{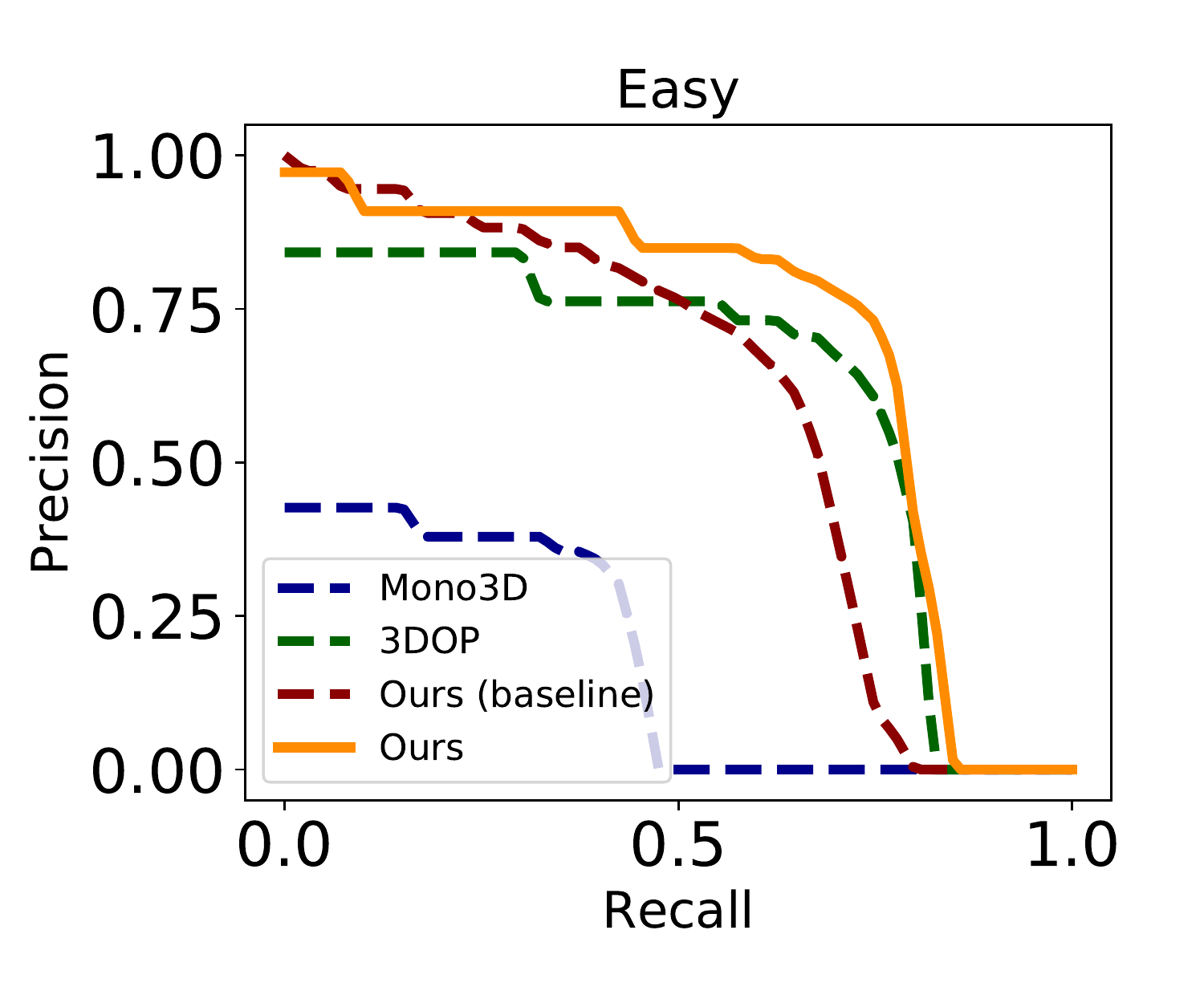} &
		\includegraphics[width=0.33\linewidth]{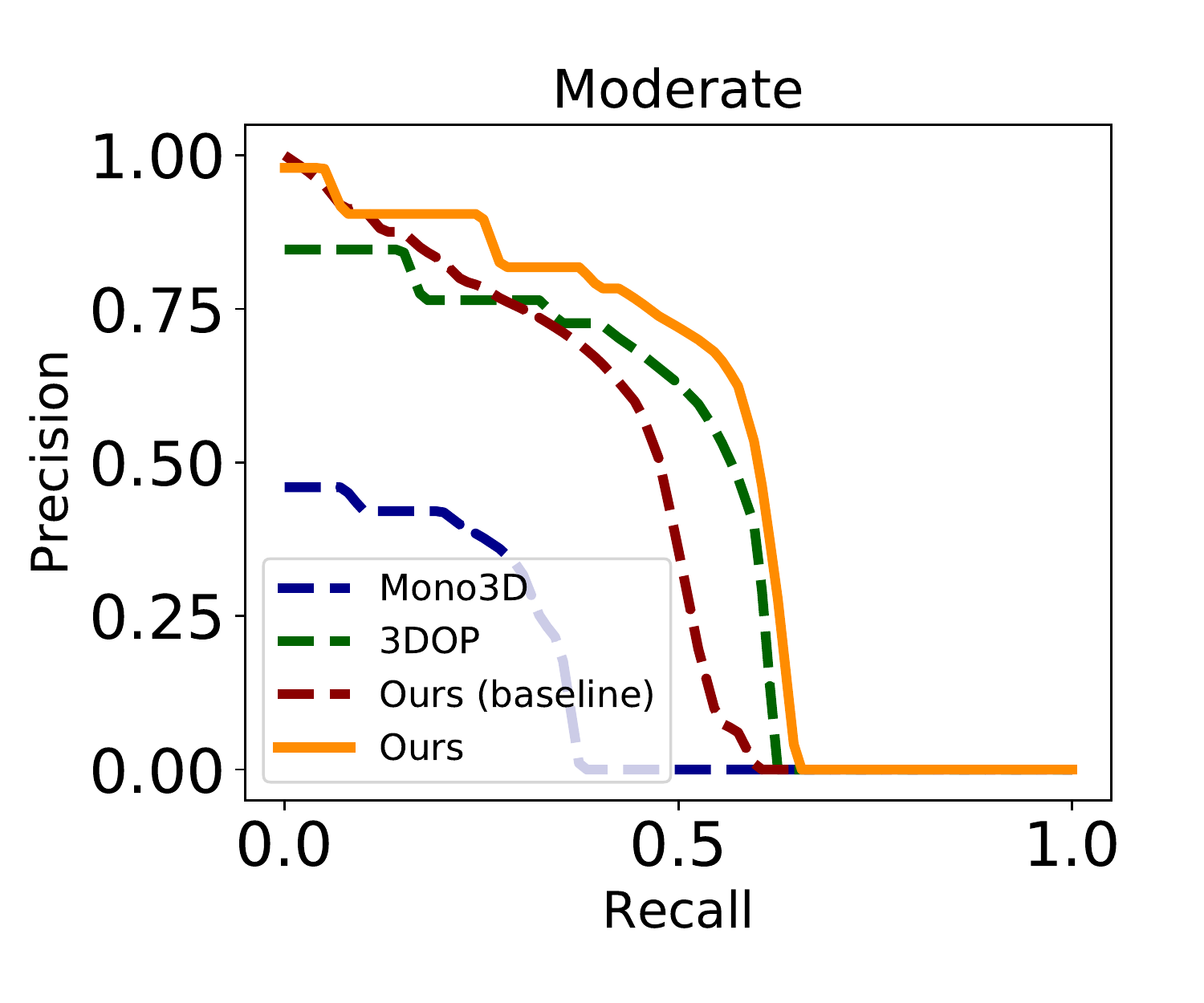} &
		\includegraphics[width=0.33\linewidth]{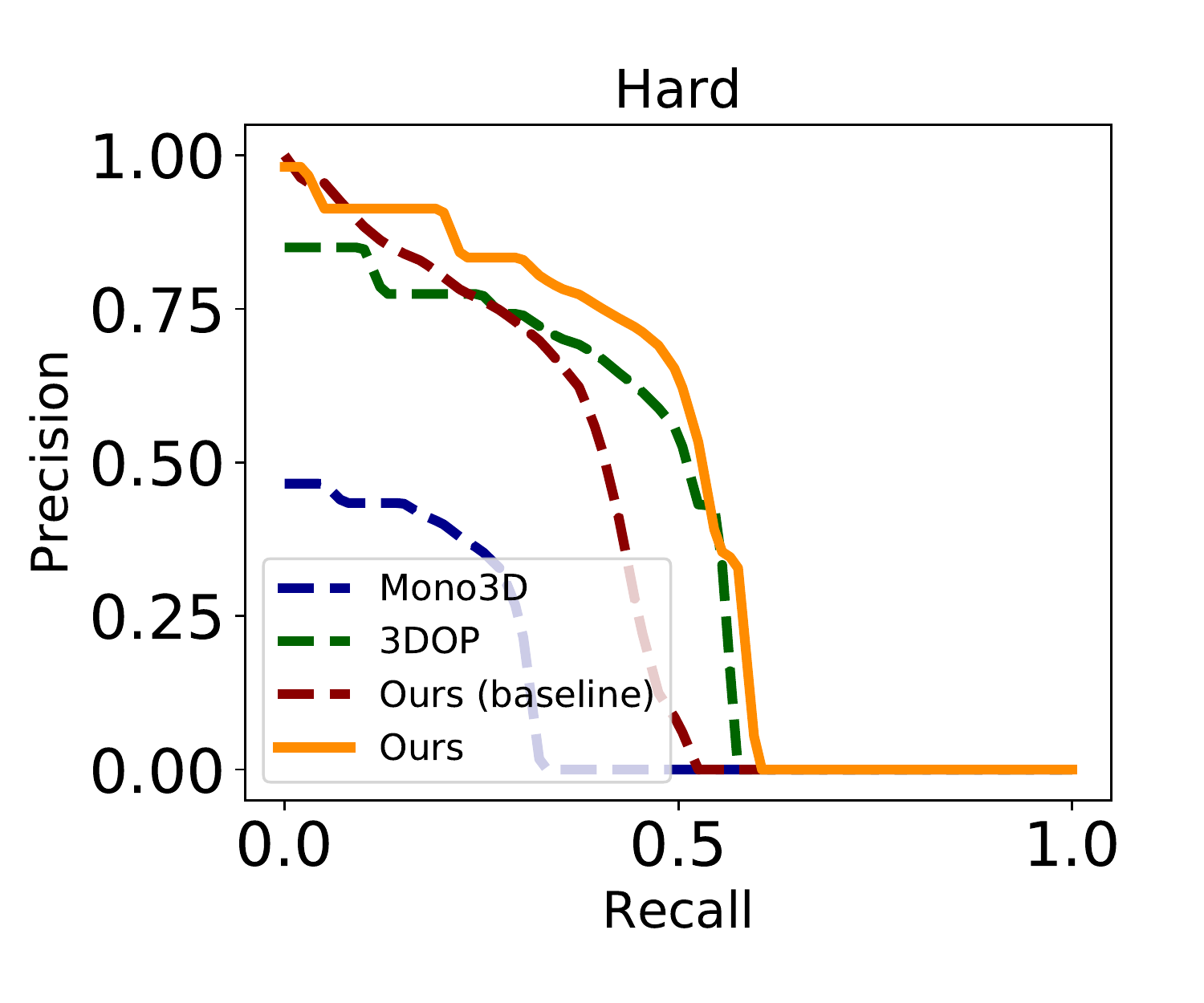} \\
		\multicolumn{3}{c}{(c) 3D localization at distance threshold of 1m. }  \\
		
	\end{tabular}
	
	\caption{\textbf{Recall-precision curves.} }
	\label{fig:apcurves}

\end{figure}

\vspace{-0.1cm}
\paragraph{TLNet.} TLNet boosts the baseline performance marginally, especially for distance threshold of 2m and 1m, where there is $\sim10\%$ gain in AP\textsubscript{LOC}. According to the Recall-Precision curves in \fig{\ref{fig:apcurves}} (c), TLNet increases both the precision and recall. The maximum precision is close to $1.0$ because most of the top predictions are correct.

\subsection{Qualitative Results}
3D bounding boxes predicted by the baseline network and our stereo method are presented in \fig{\ref{fig:visualize3d}}. In general, the predicted orange bounding boxes matches the green ground truths better when TLNet is integrated into the baseline model. As shown in (a) and (c), our method can reduce depth error, especially when the targets are far away from the camera. Object targets missed by the baseline in (b) and (f) are successfully detected. The heavily truncated car in the right-bottom of (d) is also detected, since the object proposals are in 3D, regardless of 2D truncation.

\begin{table}
	\centering
	\setlength{\tabcolsep}{3mm}{  
	    \begin{spacing}{1.1}
			\scalebox{0.8}{
			
				\begin{tabular}{|p{0.5cm}<{\centering}|p{1.6cm}<{\centering} |ccc|}
					\hline
					\multirow{2}{*}{IoU}            &\multirow{2}{*}{Method}                 & \multicolumn{3}{c|}{AP\textsubscript{3D}}                 \\ \cline{3-5}
					                                             &                                                        & \multicolumn{1}{c|}{Easy}     & \multicolumn{1}{c|}{Moderate}  & Hard    \\ \hline \hline
					                                          
                  \multirow{3}{*}{0.3}             &  Concat.                                          &  76.87                     & 60.81                          & 54.70                       \\          
                 
                                                                &  Add                                        & 75.74                         & 59.89                          & 53.57                     \\
                                                                &  Reweight                                             & \textbf{78.26}                & \textbf{63.36}                 & \textbf{57.10}         \\ \hline
                                                                
                  \multirow{3}{*}{0.5}             &  Concat.                                          & 56.32                        &41.20                           &36.41                \\          
                 
                                                                &  Add                                         &53.41                      & 41.61                         & 36.37                \\
                                                                &  Reweight                                              &\textbf{59.51}             & \textbf{43.71}                & \textbf{37.99}      \\ \hline
                                                                
                 \multirow{3}{*}{0.7}              &  Concat.                                          & 13.97                         &11.63                           &9.67                \\          
                 
                                                                &  Add                                       & 16.60                     & 13.59                         & 11.20             \\
                                                                &  Reweight                                             & \textbf{18.15}            & \textbf{14.26}                & \textbf{13.72}         \\ \hline 
				\end{tabular}  
				
			} 
			\end{spacing}
     }
	\caption{\textbf{Effect of reweighting on AP\textsubscript{3D}.} Three fusion methods in \fig{\ref{fig:fusion}} are compared, including concatenation, direct addition and the proposed reweighting strategy.}
	\label{tab:3dapnorew} 
\end{table}

\begin{figure}
    \centering
    \includegraphics[width=1\linewidth]{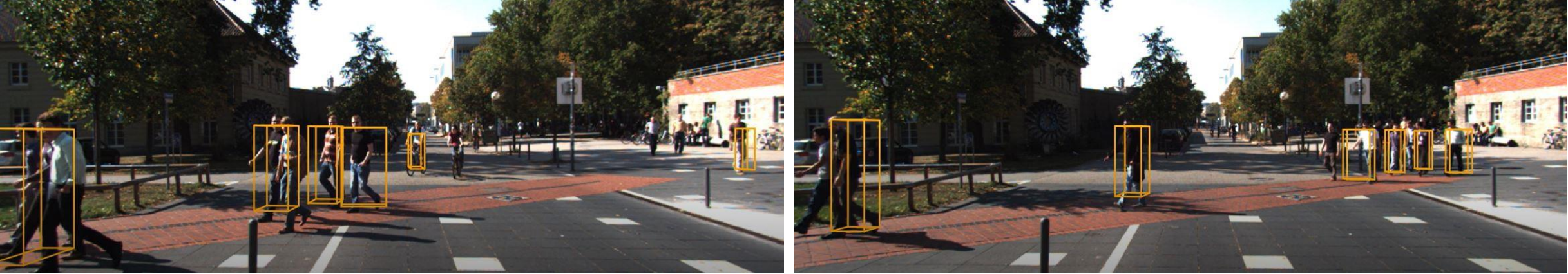}
    \caption{\textbf{Qualitative results for persons.}}
    \label{fig:vis_persons}
\end{figure}

\subsection{Ablation Study}
\begin{figure}[!htbp]
	\centering
	\includegraphics[width=1\linewidth]{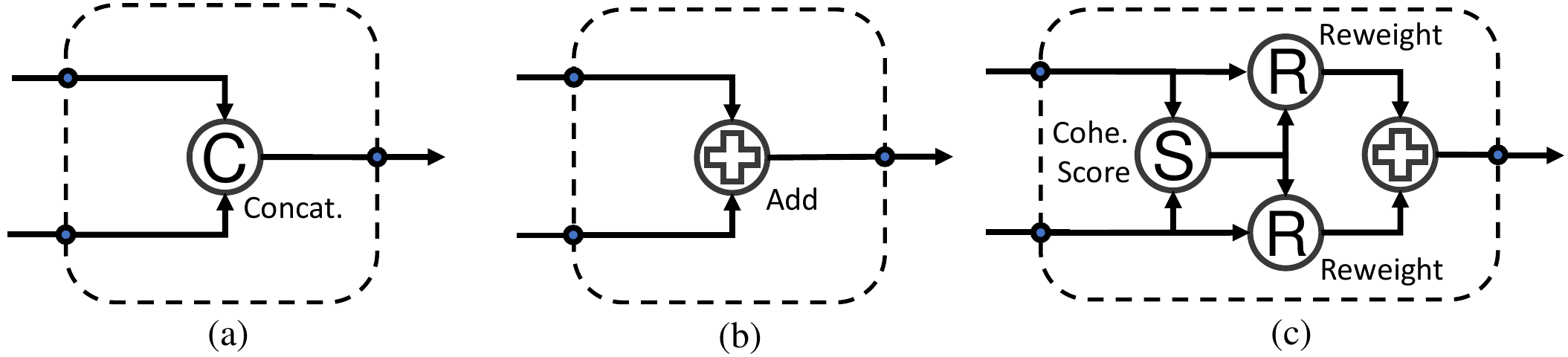}
	\caption{\textbf{Feature fusion methods in TLNet.} (c) is the proposed strategy. It first computes the coherence score of each pair of channels, then reweights the channels and adds them together.}
	\label{fig:fusion}
\end{figure}

In TLNet, the small feature maps obtained by use of RoIAlign~\cite{he2017mrcn} are not directly fused. In order to focus more attention on the key parts of an object and reduce noisy signals, we first calculate the pairwise cosine similarity $cos_i$ as coherence score, then reweight corresponding channels by multiplying with $s_i$. See \ref{subsubsec:tlnet} and \fig{\ref{fig:metriclearning}} for details. In the followings, we examine the effectiveness of reweighting by replacing it with other two fusion methods, \ie, direct concatenation and element-wise addition, as is shown in \fig{\ref{fig:fusion}}. 

Comparisons between their AP\textsubscript{3D} and AP\textsubscript{BEV} are presented in \tab{\ref{tab:3dapnorew}}. The evaluation corresponds with the empirical analysis in \ref{subsubsec:tlnet}. Since the left and right branches share their weights in the backbone network, the same channels in left and right RoIs are expected to extract the same kind of features. Some of these features are more suitable for efficient triangulation learning, which are strengthened by reweighting. Note that the coherence score $s_i$ is not fixed for each channel, but is dynamically determined by the RoI pairs cropped out at specific locations.

%% file: cvpr19_stereo.bbl
\begin{thebibliography}{10}\itemsep=-1pt

\bibitem{cai2013fastdetection}
H.~Cai, T.~Werner, and J.~Matas.
\newblock Fast detection of multiple textureless 3-d objects.
\newblock In {\em Computer Vision Systems}, pages 103--112, 2013.

\bibitem{chen2016monocular}
X.~Chen, K.~Kundu, Z.~Zhang, H.~Ma, S.~Fidler, and R.~Urtasun.
\newblock Monocular 3d object detection for autonomous driving.
\newblock In {\em Conference on Computer Vision and Pattern Recognition
  (CVPR)}, pages 2147--2156, 2016.

\bibitem{chen20153dop}
X.~Chen, K.~Kundu, Y.~Zhu, A.~G. Berneshawi, H.~Ma, S.~Fidler, and R.~Urtasun.
\newblock 3d object proposals for accurate object class detection.
\newblock In {\em Advances in Neural Information Processing Systems}, pages
  424--432, 2015.

\bibitem{chen2017multiview}
X.~Chen, H.~Ma, J.~Wan, B.~Li, and T.~Xia.
\newblock Multi-view 3d object detection network for autonomous driving.
\newblock In {\em IEEE CVPR}, volume~1, page~3, 2017.

\bibitem{cheng2018depth}
X.~Cheng, P.~Wang, and R.~Yang.
\newblock Depth estimation via affinity learned with convolutional spatial
  propagation network.
\newblock In {\em European Conference on Computer Vision}, pages 108--125,
  2018.

\bibitem{dai2018triangulation}
W.~Dai, Y.~Zhang, P.~Li, and Z.~Fang.
\newblock Rgb-d slam in dynamic environments using points correlations.
\newblock {\em arXiv preprint arXiv:1811.03217}, 2018.

\bibitem{engelcke2017vote3d}
M.~Engelcke, D.~Rao, D.~Z. Wang, C.~H. Tong, and I.~Posner.
\newblock Vote3deep: Fast object detection in 3d point clouds using efficient
  convolutional neural networks.
\newblock In {\em 2017 IEEE International Conference on Robotics and Automation
  (ICRA)}, 2017.

\bibitem{geiger2012kitti}
A.~Geiger, P.~Lenz, and R.~Urtasun.
\newblock Are we ready for autonomous driving? the kitti vision benchmark
  suite.
\newblock In {\em Computer Vision and Pattern Recognition (CVPR)}, pages
  3354--3361. IEEE, 2012.

\bibitem{xavier2010init}
X.~Glorot and Y.~Bengio.
\newblock Understanding the difficulty of training deep feedforward neural
  networks.
\newblock In {\em Aistats}, 2010.

\bibitem{hartley2003multiple}
R.~Hartley and A.~Zisserman.
\newblock {\em Multiple view geometry in computer vision}.
\newblock Cambridge university press, 2003.

\bibitem{he2017mrcn}
K.~He, G.~Gkioxari, P.~Doll{'a}r, and R.~Girshick.
\newblock Mask r-cnn.
\newblock {\em arXiv preprint arXiv:1703.06870}, 2017.

\bibitem{herrera2014slam}
C.~D. Herrera, K.~Kim, J.~Kannala, K.~Pulli, and J.~Heikkilä.
\newblock Dt-slam: Deferred triangulation for robust slam.
\newblock In {\em 2014 2nd International Conference on 3D Vision}, pages
  609--616, 2014.

\bibitem{hu2018senet}
J.~Hu, L.~Shen, and G.~Sun.
\newblock Squeeze-and-excitation networks.
\newblock 2018.

\bibitem{kehl2017ssd6d}
W.~Kehl, F.~Manhardt, F.~Tombari, S.~Ilic, and N.~Navab.
\newblock Ssd-6d: Making rgb-based 3d detection and 6d pose estimation great
  again.
\newblock In {\em Proceedings of the International Conference on Computer
  Vision (ICCV 2017)}, pages 22--29, 2017.

\bibitem{kim2018lidar}
J.-U. Kim and H.-B. Kang.
\newblock A new 3d object pose detection method using lidar shape set.
\newblock {\em Sensors}, 18(3):882, 2018.

\bibitem{kingma:2014}
D.~P. Kingma and J.~Ba.
\newblock Adam: A method for stochastic optimization.
\newblock In {\em International Conference for Learning Representations}, 2014.

\bibitem{ku2017joint}
J.~Ku, M.~Mozifian, J.~Lee, A.~Harakeh, and S.~Waslander.
\newblock Joint 3d proposal generation and object detection from view
  aggregation.
\newblock {\em arXiv preprint arXiv:1712.02294}, 2017.

\bibitem{lahond2017twoddriven}
J.~Lahoud and B.~Ghanem.
\newblock 2d-driven 3d object detection in rgb-d images.
\newblock In {\em 2017 IEEE International Conference on Computer Vision
  (ICCV)}, 2017.

\bibitem{li20173dfcn}
B.~Li.
\newblock 3d fully convolutional network for vehicle detection in point cloud.
\newblock In {\em 2017 IEEE/RSJ International Conference on Intelligent Robots
  and Systems (IROS)}, 2017.

\bibitem{li2016}
B.~Li, T.~Zhang, and T.~Xia.
\newblock Vehicle detection from 3d lidar using fully convolutional network.
\newblock 2016.

\bibitem{lin2016fpn}
T.-Y. Lin, P.~Doll{'a}r, R.~Girshick, K.~He, B.~Hariharan, and S.~Belongie.
\newblock Feature pyramid networks for object detection.
\newblock {\em arXiv preprint arXiv:1612.03144v2}, 2017.

\bibitem{mahjourian2018unsupervised}
R.~Mahjourian, M.~Wicke, and A.~Angelova.
\newblock Unsupervised learning of depth and ego-motion from monocular video
  using 3d geometric constraints.
\newblock In {\em Proceedings of the IEEE Conference on Computer Vision and
  Pattern Recognition}, pages 5667--5675, 2018.

\bibitem{matthew2014vgg}
D.~Z. Matthew and F.~Rob.
\newblock Visualizing and understanding convolutional networks.
\newblock In {\em European Conference on Computer Vision}, pages 818--833,
  2014.

\bibitem{mayer2016dispnet}
N.~Mayer, E.~Ilg, P.~Hausser, and P.~Fischer.
\newblock A large dataset to train convolutional networks for disparity,
  optical flow, and scene flow estimation.
\newblock In {\em Computer Vision and Pattern Recognition(CVPR)}, pages
  4040--4047, 2016.

\bibitem{nir2004causal}
T.~Nir and A.~M. Bruckstein.
\newblock Causal camera motion estimation by condensation and robust statistics
  distance measures.
\newblock In {\em European Conference on Computer Vision (ECCV)}, pages
  119--131, 2004.

\bibitem{petroff2018triangulation}
E.~Petroff, L.~C. Oostrum, B.~W. Stappers, M.~Bailes, E.~D. Barr, S.~Bates,
  S.~Bhandari, N.~D.~R. Bhat, M.~Burgay, S.~Burke-Spolaor, A.~D. Cameron, D.~J.
  Champion, R.~P. Eatough, C.~M.~L. Flynn, A.~Jameson, S.~Johnston, E.~F.
  Keane, M.~J. Keith, L.~Levin, V.~Morello, C.~Ng, A.~Possenti, V.~Ravi, W.~van
  Straten, D.~Thornton, and C.~Tiburzi.
\newblock A fast radio burst with a low dispersion measure.
\newblock {\em arXiv preprint arXiv:1810.10773}, 2018.

\bibitem{qin2019monogr}
Z.~Qin, J.~Wang, and Y.~Lu.
\newblock Monogrnet: {A} geometric reasoning network for monocular 3d object
  localization.
\newblock {\em AAAI}, 2019.

\bibitem{ren2017faster}
S.~Ren, K.~He, R.~Girshick, and J.~Sun.
\newblock Faster r-cnn: towards real-time object detection with region proposal
  networks.
\newblock {\em IEEE Transactions on Pattern Analysis \& Machine Intelligence},
  2017.

\bibitem{ren2016layout}
Z.~Ren and E.~B. Sudderth.
\newblock Three-dimensional object detection and layout prediction using clouds
  of oriented gradients.
\newblock In {\em 2016 IEEE Conference on Computer Vision and Pattern
  Recognition (CVPR)}, 2016.

\bibitem{xu2018multifusion}
B.~Xu and Z.~Chen.
\newblock Multi-level fusion based 3d object detection from monocular images.
\newblock In {\em Computer Vision and Pattern Recognition (CVPR)}, pages
  2345--2353, 2018.

\bibitem{zbontar2016stereo}
J.~Zbontar and Y.~LeCun.
\newblock Stereo matching by training a convolutional neural network to compare
  image patches.
\newblock {\em Journal of Machine Learning Research}, 17(1-32):2, 2016.

\bibitem{zeng2017multiview}
A.~Zeng, K.-T. Yu, S.~Song, D.~Suo, E.~Walker, A.~Rodriguez, and J.~Xiao.
\newblock Multi-view self-supervised deep learning for 6d pose estimation in
  the amazon picking challenge.
\newblock In {\em 2017 IEEE International Conference on Robotics and Automation
  (ICRA)}, 2017.

\end{thebibliography}
